\documentclass[letterpaper, 10 pt, conference]{ieeeconf}  

\usepackage{xcolor}
\usepackage{cite}
\usepackage{graphicx}
\usepackage{subcaption}
\usepackage{booktabs} 
\usepackage{gensymb}
\usepackage{array} 
\usepackage{pifont}
\usepackage{xspace}
\usepackage{hyperref}
\hypersetup{%
  colorlinks=true,%
  linkcolor={red!50!black},
  citecolor={blue!65!black},
  urlcolor={blue!65!}
  }

\title{\LARGE \bf
ContactHandover: \\Contact-Guided Robot-to-Human Object Handover
}

\author{Zixi Wang$^{1}$ \quad Zeyi Liu$^{2}$ \quad Nicolas Ouporov$^{1}$ \quad Shuran Song$^{1,2}$ \\
$^{1}$Columbia University \quad \quad $^{2}$Stanford University 
\\
}

\newcommand\name{ContactHandover\xspace}
\newcommand\GF{Grasp Re-ranking \xspace}
\newcommand\HO{Handover Orientation\xspace}

\begin{document}

\maketitle

\thispagestyle{empty}
\pagestyle{empty}

\begin{abstract}
Robot-to-human object handover is an important step in many human robot collaboration tasks. A successful handover requires the robot to maintain a stable grasp on the object while making sure the human receives the object in a natural and easy-to-use manner. We propose \name, a robot to human handover system that consists of two phases: a contact-guided grasping phase and an object delivery phase. During the grasping phase, \name predicts both 6-DoF robot grasp poses and a 3D affordance map of human contact points on the object. The robot grasp poses are re-ranked by penalizing those that block human contact points, and the robot executes the highest ranking grasp. During the delivery phase, the robot end effector pose is computed by maximizing human contact points close to the human while minimizing the human arm joint torques and displacements. We evaluate our system on 27 diverse household objects and show that our system achieves better visibility and reachability of human contacts to the receiver compared to several baselines. 
More results can be found on the \href{https://clairezixiwang.github.io/ContactHandover.github.io/}{project website}.

\end{abstract}

\section{INTRODUCTION}
Object handover is a key step towards natural human and robot collaboration~\cite{human_pref_cakmak, affordance_mobility, robot_pass_tool, castro2021trend, ortenzi2021object}. Robot-to-human handover, in particular, has wide applications in a lot of practice scenarios, from handing tools to workers in factory to fetching daily objects for elders at home. A handover process typically involves a grasping phase in which the robot picks up the target object, and a delivery phase where the robot approaches the human and moves the object to a pose that is accessible and ergonomic for the human receiver to grasp and use in subsequent tasks.

There are two key challenges in performing a successful robot to human handover: first, when grasping the object to be handed over, the robot needs to leave room for the human receiver to grasp the object while also choosing a stable grasp pose.
For example, the robot should choose a stable grasp on the head of the hammer and leave enough room on the handle of the hammer for the receiver.
Second, the robot should deliver the object in a way that most natural grasping areas are visible and reachable from the receiver. For example, the robot should orient the handle of the hammer to the receiver instead of the head.

To address these challenges, we propose a robot-to-human handover system, \textbf{\name}, which uses 3D contact maps to model diverse human preferences when receiving objects. Our system contains two phases. In the object grasping phase, the system predicts 6-DoF robot grasp poses and a human contact map for the object, re-ranks the grasp poses to penalize those that occlude human contacts, and executes the highest ranking grasp. During the delivery phase, we compute the robot end-effector position and orientation that both minimizes the human arm joint torques and displacements when receiving the object, as well as the distances from contact points to the receiver eyes' location.

To evaluate our result quantitatively, we propose two computational metrics, visibility and reachability, that align with previous work's discovery on ergonomic object delivery poses for human receivers~\cite{robot_pass_tool, affordance_mobility}. The visibility metric measures the percentage of human contact points from the receiver's viewpoint. The reachability metric measures the percentage of human contact points reachable from the receiver without obstruction from the robot's embodiment. Finally, a handover is considered "successful" if both the visibility and reachability metric exceed a threshold.

\begin{figure}[t!]
    \centering
    \includegraphics[width=0.98\linewidth]{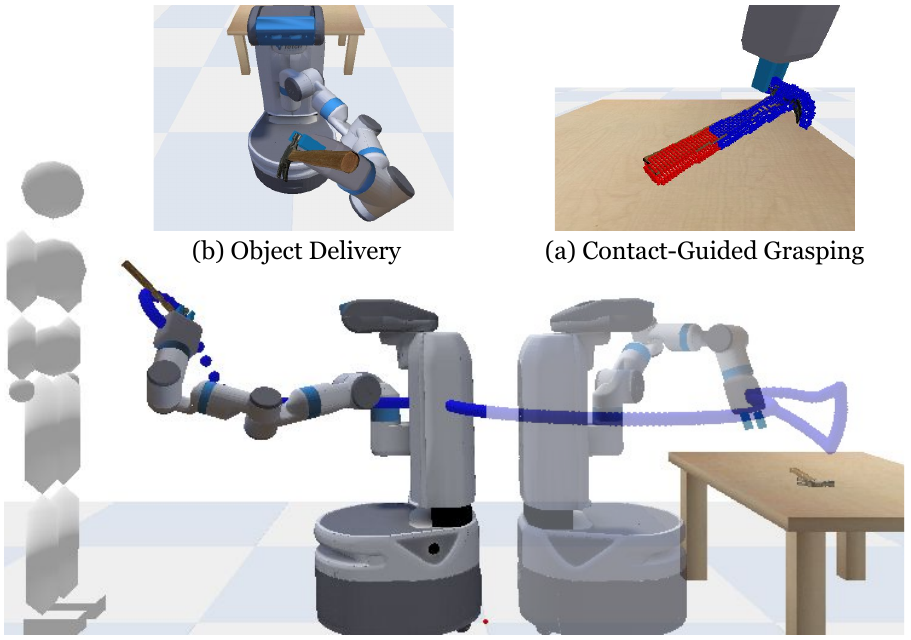}
    \caption{\small\textbf{Contact-Guided Robot to Human Object Handover.} We propose a robot-to-human handover system with two phases: (a) contact-guided grasping and (b) object delivery. (a) During grasping, the robot predicts 6-DoF grasp poses and human contact points (denoted in red) for the object, and selects a grasp pose that maximizes stability while minimizing contact points occlusions. (b) During delivery, the robot computes a handover location and orientation that minimizes human arm joint torque and displacements, as well as the distance between contact points and the human.}
    \label{fig:teaser}
\vspace{-3mm}
\end{figure}

In summary, our main contribution is a robot to human handover system that maximizes the visibility and reachability of human preferred contact points to the human receiver. To achieve this, we introduce:
\begin{itemize}
    \item A contact-guided grasp selection algorithm that accounts for both grasp stability of the robot and contact preferences of the human receiver.
    \item An object delivery algorithm that computes the robot end effector pose by considering the human's arm comfort when receiving the object and minimizing the distance between the contact points and the human.
    \item Two benchmark metrics (visibility, reachability) that quantitatively evaluate a handover pose.
\end{itemize}
\vspace{-1mm}
\section{RELATED WORK}
\textbf{Grasp Pose Prediction} has been a long standing task in robotics.
Early works use analytical methods to plan and execute stable grasps, which are limited in real world applications given the assumption of known object geometry~\cite{ferrari1992planning, miller2003automatic, goldfeder2007grasp}. 
More recently, data-driven approaches use convolutional neural networks to learn grasp affordances directly from top-down RGB-D images~\cite{mahler2017dex,zeng2022robotic, he2023pick2place}, or use generative models such as VAE to sample grasp candidates from object point cloud and then filter based on the grasping quality \cite{mousavian20196, sundermeyer2021contact}. In addition to predicting stable grasps, other works have studied predicting functional grasps, namely grasping on functional parts of the object to use~\cite{brahmbhatt2019contactgrasp, 2023functional, geng2023rlafford, zhu2023toward, liu2023busybot}.
In the robot-to-human handover context, we identify functional grasps as those that avoid human preferred contact points, and propose a contact-guided grasp ranking method to predict stable grasp that also maximizes available human contact points on the object.
\vspace{1mm}

\textbf{Learning Human Grasp Affordances. }
During a robot-to-human handover, it is important to generate grasp pose and delivery pose that accommodate human grasp affordances. To model human grasp affordances, one line of works directly learn 3D affordance maps on objects with respect to different intents (e.g. use, handoff)~\cite{contactdb,ardon2019learning, deng20213d}. In particular, ContactDB~\cite{contactdb} is a dataset of human contact maps for household objects collected with a thermal camera. Other works model human grasp affordance by predicting hand shape and pose when grasping ~\cite{corona2020ganhand,jiang2021hand,ye2021h2o,ye2023affordance,fan2023hold}. In this work, we use 3D human contact affordance maps as a proxy for human preferences while receiving objects and show that the contact maps can be used to effectively guide robot grasp pose and delivery pose selection during a robot-to-human handover.

\vspace{1mm}
\textbf{Robot-to-Human Object Handover. }
The main objective of robot-to-human object handover is to deliver the object in a way that maximizes the user's ease to grasp and convenience to use the object for a subsequent task \cite{10.3389/frobt.2020.542406}. Prior works have attempted to predict and maximize human grasp affordance during handover, but they either manually select human grasping part on an object~\cite{Aleotti2014-xj},  make assumptions about object geometry and hand-design object categories~\cite{determine_grasp_config, affordance_mobility, kang2023safe}, or train on synthetic data that fail to model the complexity of human contacts~\cite{handover-affordance}. Other works use simplifying heuristics such as assuming that the robot grasp is on the opposite side of a predicted human hand pose~\cite{fast_and_comfortable}. But this heuristic does not account for scenarios where both the human and robot have to grasp on same side of an object (e.g. knife handle). We propose a novel approach that predicts human contact points on an object with no assumption of object category and use human contact points to guide robot grasp pose and delivery pose selection.

\begin{figure*}[ht!]
    \centering
    \includegraphics[width=0.95\linewidth]{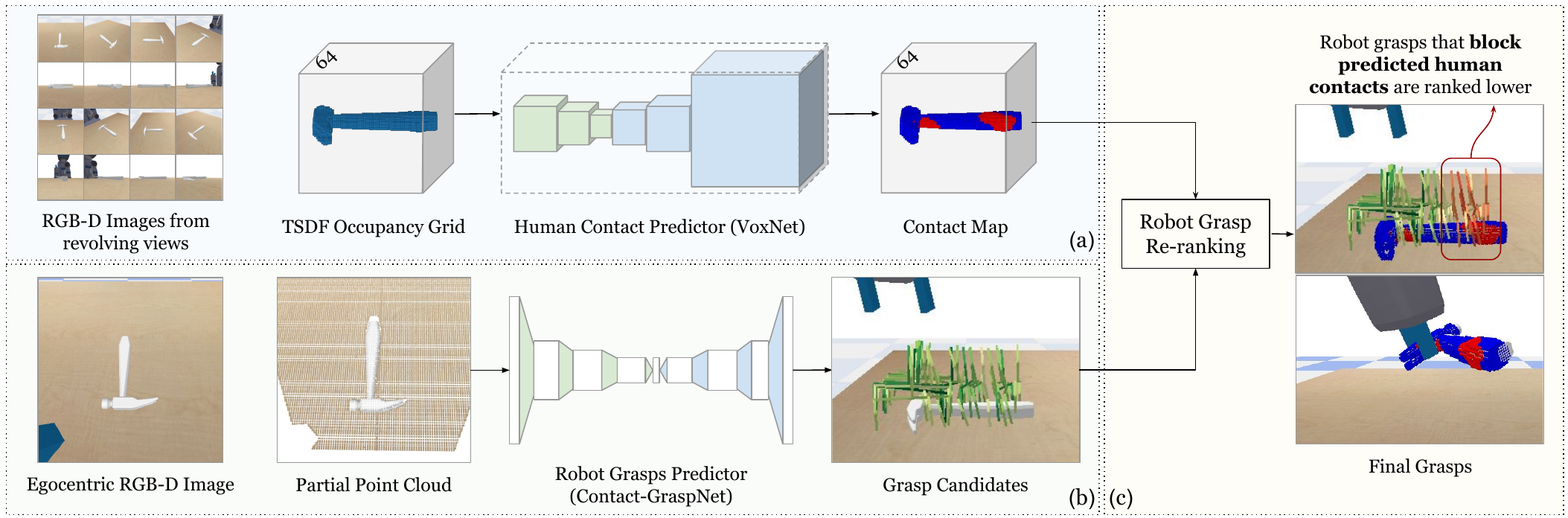}
    
    \caption{\textbf{Contact-Guided Grasp Selection.} (a)\S \ref{method:human_contact} The robot takes RGB-D Images from 16 views around the table and construct a $64^3$ voxel representation of the object via TSDF fusion; the occupancy grid is then fed into a trained 3D VoxNet to predict human contact maps.  (b)\S \ref{method:robot_grasp} The robot takes a partial point cloud observation as input to the pre-trained Contact-GraspNet model to generate a set of 6-DoF robot grasps. (c)\S \ref{method:GF} The robot executes the grasp with highest score as computed by Equation \ref{eq:grasp_filter}.}
    \label{fig: Grasping}
\end{figure*}
\section{METHOD}
\name contains two phases: a contact-guided grasping phase and an object delivery phase. During the grasping phase, given RGB-D observations of an object on table, the system predicts both 6-DoF robot grasp pose candidates and human contact points on the object, and then re-ranks the grasp poses by penalizing grasps poses based on number of human contact points occluded by the robot end effector. During the delivery phase, the system computes a handover position and orientation that minimizes the human arm joint torques and displacements, and minimizes the total distances from contact points to the receiver’s eyes location. We will discuss each module in detail below.

\vspace{-0.5mm}
\subsection{\textbf{Contact-Guided Grasp Selection}}
A successful handover requires the robot to select a stable grasp location on the object that leaves room for the human receiver's grasp later.
As illustrated in Fig. \ref{fig: Grasping}, given RGB-D observations of an object on the table, the robot predicts a set of grasp pose candidates together with a human contact map on the object. A final score is computed for each robot grasp pose candidate by combining the confidence score of the grasp and percentage of occluded human contact points. The grasp with the highest score is chosen and executed.

\subsubsection{\textbf{Human Contact Prediction}}
\label{method:human_contact}

We leverage the ContactDB ``use'' dataset which contains 27 objects and 50 contact maps (collected from 50 participants) for each object. Each contact map is a $64^3$ voxel grid, where a voxel is labeled as 1, if contacted by human during the grasp, or 0, if not contacted. We randomly select one contact map for each object during training.

\underline{Human Contact Model:}
We train a 3D VoxNet\cite{voxnet} on the contactDB ``use'' dataset. The model takes in a solid occupacy grid of the object in a $64^3$ voxel space and predicts whether each voxel will be contacted during a human grasp. Following \cite{contactdb}, we enforce cross entropy loss only on the voxels on the object surface.

\underline{Predicting Human Contacts:}
We record RGB-D observations from 16 views around the table and use TSDF fusion~\cite{zeng20163dmatch} to construct a $64^3$ voxel grid. The voxel grid is then fed to the human contact model to predict a human contact map on the object surface voxels.

\subsubsection{\textbf{Robot Grasp Prediction}}
\label{method:robot_grasp}
We want to generate a diverse set of robot grasp pose candidates to increase the likelihood of attaining stable grasps while minimizing number of human contacts blocked by robot end effector. To do so, we use the pre-trained Contact-GraspNet model~\cite{sundermeyer2021contact}.

Contact-GraspNet is a PointNet++ based U-shaped model that takes in a partial point cloud observation of a scene, and for each point $i$, predicts whether it is contacted by the robot gripper during grasping with a confidence score $S(i)$. For each point $i$, the model predicts the 3-DoF grasp orientation and grasp width $w \in R$ of a parallel-yaw gripper. This 4-DOF grasp representation can then be translated to a 6-DoF robot gripper pose $g$ for each contact point $i$. Following \cite{sundermeyer2021contact}, we select grasps with confidence $S(g)\geq0.23$ as robot grasp candidates. We use the predicted confidence score for each grasp $S(g)$ in the re-ranking phase.

\subsubsection{\textbf{Robot Grasp Re-ranking}}
\label{method:GF}
Given both human contact points and robot grasp candidates, we select the grasp pose that has a high contact confidence score and minimizes the number of human contact points blocked by the robot end effector. To do this, we re-score the robot grasp pose candidates by penalizing human contact occlusions. We cluster the predicted contact points and minimize human contact occlusion for the biggest cluster.

\underline{Clustering Human Contacts:}
For some objects, like the binoculars, people typically use them with both hands. However, during handovers, a person will likely receive these objects with one hand. Moreover, avoiding all possible contact points might leave few valid robot grasp candidates after re-ranking, leading to grasp failures. To resolve this, we cluster predicted human contact points using DBSCAN\cite{dbscan} based on point spatial density, with no assumptions on the number of clusters. We then choose the largest human contact cluster to compute human contact occlusion and handover orientation in \S \ref{method:HO}. 

\underline{Grasp Re-Ranking:}
For each grasp $g$, we compute the percentage of human contact points that are occluded by the robot gripper, denoted as $O(g)$. For each human contact point, we ray trace from the contact point along it's surface normal and check if the ray collides with the robot gripper at grasp pose $g$. If there's a collision, then we consider the human contact point to be blocked by the gripper at pose $g$. Specifically, $O(g)$ is computed as
\begin{equation}
    O(g) = \frac{\Big|\big\{i\in C_{pred} \mid i \textrm{ blocked by } g  \big\}\Big|}{\big|C_{pred}\big|},
\end{equation}
where $C_{pred}$ is the largest cluster of predicted contact points on the object surface.

Lastly, We re-rank the robot grasps to account for the number of human contacts they occlude. We compute a final contact score $C(g)$ for each grasp, which combines the grasping contact confidence $S(g)$ and the occluded human contacts $O(g)$:
\begin{equation}
\label{eq:grasp_filter}
    C(g) = \lambda S(g) - (1-\lambda)O(g).
\end{equation}
$\lambda$ controls the weight between grasp confidence and human contact occlusion. We use $\lambda=0.5$ in our experiments.
Finally, the robot executes the grasp with the highest $C(g)$.

\vspace{-0.5mm}
\subsection{\textbf{Handover Position}}
During the delivery phase, the robot should hand the object to a point in front of the human that is both reachable and comfortable for the human arm, with respect to the human's height and pose. For instance, the handover position for a human that is standing should differ from a human that is sitting down.
Following previous works \cite{Katayama2003_arm_pose_comfort, Parastegari2017_modeling_human_2d, Liu2021_HumanComfort_3d}, \name computes the point of handover by minimizing human arm joint torques and joint displacement. In addition, we enforce the handover position to be below the human's shoulder and above the waist.

We estimate the point of handover with respect to the human's shoulder location, in order to account for human height and pose variety. We assume access to the ground truth human shoulder location together with upper and lower arm lengths. Following \cite{Parastegari2017_modeling_human_2d}, we also assume the receiver's reaching motion trajectory lies on the vertical plane of the human receiver's right arm, and estimate the handover location on this vertical plane. 

\underline{Joint Torques.}
For each point $(x, y, z)$ in space, we compute the total joint torque of the human arm to hold an object at that point, where $n$ represents the number of joints, $\tau_j$ represents the torque of the joint $j$, $c_{t,max}$ represents the maximum cost value of all points.
\begin{equation}
  f_{torque}(x, y, z) = \frac{\sum_{j=1}^n (\tau_j)^2}{c_{t,max}}
\end{equation}

\underline{Joint Displacements.}
For each point $(x, y, z)$ in space, and for each human arm configuration to reach that position, we calculate how far each joint deviates from the medium angle of its range of motion:
\begin{equation}
    f_{disp}(x,y,z)=\frac{\sum_{j=1}^n (\theta_{mid,j} - \theta_{j})^2}{c_{d,max}}
\end{equation}
where $\theta_{mid,j}$ is the medium value of the angle range of the joint $j$,
$\theta_j$ is the rotation angle of joint $j$, $c_{d,max}$ is the maximum cost value. Based on the study in \cite{Liu2021_HumanComfort_3d}, 
the medium angle for the shoulder's forward-backward rotation is $67.5\degree$, with respect to the torso; and the medium angle for the elbow's forward-backward rotation is $62.5\degree$, where a straight arm pointing forward is $0\degree$, and the medium angle for bending the elbow close to the upper arm is $140\degree$.\\

Finally, we define the total cost of a candidate point as
\begin{equation}
  f_{total}(x,y,z) = ( 1 - \alpha ) f_{torque}(x,y,z) + \alpha f_{disp}(x,y,z)  
\end{equation}
where $\alpha$ controls the weight of the two cost functions.

We sample candidate handover positions by searching through the human's shoulder and elbow angles with a $5 \degree$ granularity, and using the resulting hand locations as candidate handover positions. We only select candidate positions that are lower than the human shoulder and above the waist. We choose the one that minimizes the total cost $f_{total}$ as the handover position.
\vspace{-0.5mm}
\subsection{\textbf{Handover Orientation}}
\label{method:HO}
In the delivery stage, the robot must present the object in an orientation that minimizes the total distances from contact points to the receiver’s eyes. Specifically, we uniformly sample in the spherical space with a 45 degree granularity and select the object orientations that are kinematically feasible \cite{fetch_reachability} for the robot arm at the computed handover point. For each candidate orientations, \name computes the total distance between predicted contact points and the human eyes. We estimate the human eye to be located in a position that is exactly 1/2 the height of the human head. The orientation that yields minimum distance is selected.

Note that as in \S \ref{method:GF}, we compute the distance for the largest human contact cluster. For objects with bimodal human contacts (e.g. binoculars), we found that orienting one cluster of human contact towards the human results in more natural handovers, since human tends to receive these objects with one hand instead of two hands. We show the qualitative result for the clustering in \S \ref{result:cluster} and Figure \ref{fig:cluster}.

\section{EVALUATION}
\label{sec:eval}

We evaluate \name's ability to hand over an object to the human receiver in a natural and ergonomic manner. We propose two computational metrics, visibility and reachability, to quantify the quality of a handover. We show that our system yields better handover results comparing to several ablations in \S \ref{sec:result}.

\vspace{-0.5mm}
\subsection{\textbf{Handover Metrics}}
\label{sec:metrics}
We evaluate our system on 27 daily objects from the ContactDB dataset, which consists of 50 contact maps collected from 50 different users for each object. In this section, we introduce two quantitative metrics based on the contact maps and define success for a robot-human handover.

\textbf{Metric 1: Human Contact Visibility.}
To ensure that humans can easily grasp their preferred contact areas upon receiving an object, it is crucial that these areas are visible to the receiver. In particular, the areas where humans prefer to make contact should not be occluded by the object itself, occluded by the robot's embodiment, or inside the gripper which would be effectively unavailable to the receiver either.

We design a metric that captures the visibility of human contact areas. Given a ground truth contact map $\mathbf{}CM$ of an object, We define the human contact visibility as the percentage of ground truth hand-object contacts that are visible from the human’s view. \\
\begin{equation}
    Visibility = \frac{\sum_{i\in \mathrm{V}} CM(i)}{\sum_{j\in \mathrm{P}} CM(j)}
\end{equation}
where $CM: i \mapsto c\in\{0,1\}$ indicates if each voxel $i$ on the object surface is touched during a human grasp. $\mathit{P}$ is the set of all voxels in the contact map. $\mathit{V}$ is the set of voxels that are visible from the human's view without occlusion from the robot embodiment or the object itself, and that are not inside the gripper. We compute visibility by setting a RGB-D camera at the human receiver's eye position looking at the object; and we consider any contact points that are captured in the image and are not inside the robot gripper to be visible.

\begin{figure*}[ht!]
    \centering
    \includegraphics[width=0.9\linewidth]{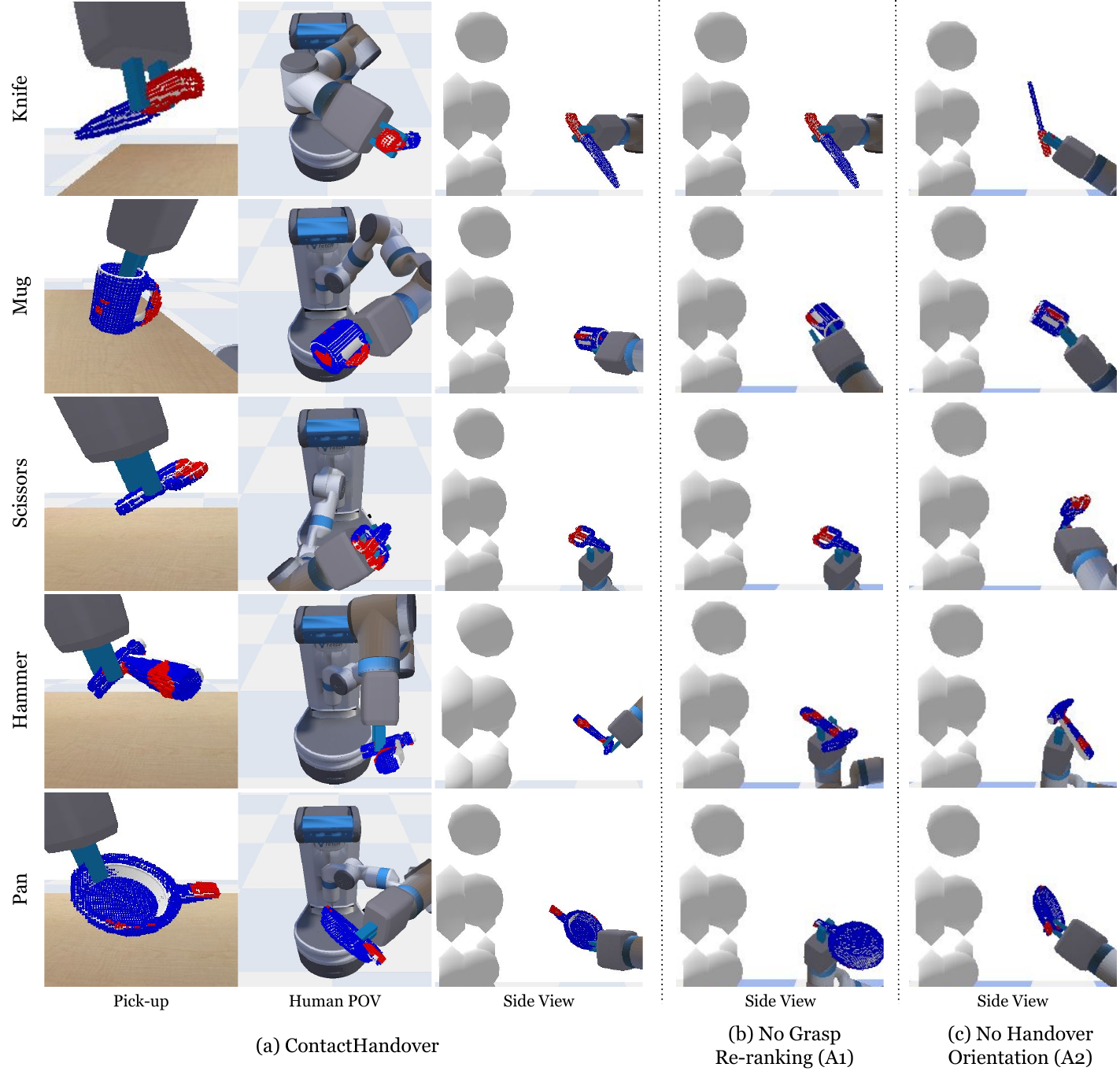}
\caption{\textbf{Qualitative Results and Ablations.} As shown in (a), \name predicts the human contact map (red indicate human contact points, and blue non-contact points), picks up the objects while avoiding human contacts, and orients the human preferred contacts towards the human during delivery. In (b), without grasp re-ranking, the robot gripper blocks human contacts, i.e., the handle of \textit{pan} and \textit{hammer}. In (c), without handover orientation, the human contacts, i.e. the handles of the \textit{scissors}, \textit{hammer}, and \textit{pan}, points away from the human. More qualitative results can be found on the \href{https://clairezixiwang.github.io/ContactHandover.github.io/}{project website}.}
    \label{fig:ablation_visuals}
\end{figure*}

\textbf{Metric 2: Human Grasp Reachability.} For a natural grasp, visibility is not sufficient by itself, human preferred contact areas should also be easily reachable by the human.
Therefore our second metric computes the percentage of human contact points that are within the human's arm reach and located in between the robot gripper and the human:
\begin{equation}
    Reachability = \frac{\sum_{i\in \mathrm{R}} CM(i)}{\sum_{j\in \mathrm{P}} CM(j)}
\end{equation}
where $\mathit{R} = \{i: (d_1(i) < l_{\mathbf{arm}}) \wedge (d_2(i) < d_2(\mathbf{gripper}))\}$. $d_1$ represents the distance from point $i$ to human shoulder and $l_{\mathbf{arm}}$ is the human's arm length. $d_2(i)$ represents the shortest horizontal distance from point $i$ to the human; $d_2(\mathbf{gripper})$ represents the shortest distance between the robot gripper and human. If  $d_2(i) < d_2(\mathbf{gripper})$, then the point $i$ is located between the human and the robot gripper. Higher reachability score indicates more human preferred contact areas are facing towards the human.

\textbf{Success. }
For each object, we compute the visibility and reachability score based on the 50 corresponding contact maps and select the median as the final score. We consider a handover to be successful if \textbf{both} visibility and reachability scores exceeds a threshold $k=0.5$.

\vspace{-1mm}
\subsection{\textbf{Experiment Setup}}
We evaluate our method in the Pybullet Simulation Environment with a Fetch Robot and a human figure that’s 1.7 meters tall.
The robot starts in front of the table, and the human is standing 2 meters behind the robot.

We set up a revolving RGB-D camera that captures 16 distinct angles around the tabletop to reconstruct the object point cloud and voxel grids. Specifically, we take two set of eight images, one at table height and another at one meter above the table; in both sets, the camera revolves around and looks at the center of the table. We also simulate a RGB-D camera on the human eye level to observe the handover object and compute the visibility metric.

For each object, the robot grasps the object on table, turns around and moves to 1.2 meters in front of the human, and finally handovers the object in the computed pose. We run the experiments for 27 selected objects from the ContactDB ``use'' dataset under 5 random seeds and report the average.

\subsection{\textbf{Baselines}}
\label{sec:baseline}
To evaluate the importance of different components in our approach, we conduct the following ablation studies as shown in Table \ref{tab:Main Results}:

\textbf{Ablation 1 (A1): No \GF (GR). } The robot does not predict or consider human contacts when grasping the object. Instead, the robot only predicts and executes the grasp with the highest confidence score.

\textbf{Ablation 2 (A2): No \HO (HO). } The robot does not calculate the object orientation during handover. Instead, the robot moves the end effector to the handover position with a random orientation.

\textbf{Ablation 3 (A3). Handover Position Only}.  The robot executes the grasp with highest confidence score, and moves the end effector to the same handover position as \name, with a random orientation.

\textbf{Ablation 4 (A4). No optimization. } The robot executes the grasp with the highest confidence score, moves the robot base to the same location as \name, while maintaining the same end effector pose after grasping.\\

\begin{figure*}[ht!]
    \centering
    \includegraphics[width=0.95\linewidth]{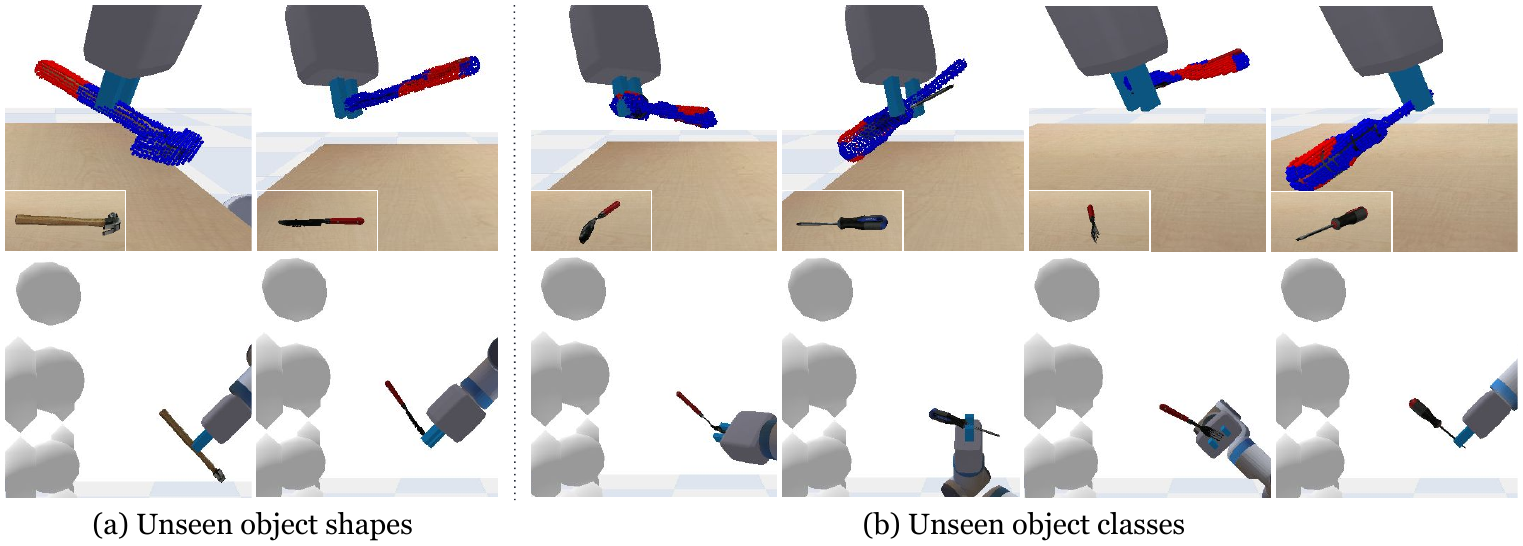}
    \caption{\textbf{Generalize to Unseen Objects.} We show \name's performance on unseen YCB objects. We show both human contact predictions and handover results on these objects. \name is able to generalize to (a) objects with unseen shapes (e.g. hammer and knife) and (b) objects with unseen types (e.g. spoon, flat screwdriver, fork and phillips screwdriver). It predicts reasonable human contacts (denoted in red points) around the handles of the objects, picks up and delivers the objects to human with respect to the predicted human contacts.}
    \label{fig:ycb}
\end{figure*}

\begin{table}[t!]
    \centering
    \begin{tabular}{cccc|c|c|c} \hline 
     & \multicolumn{3}{c|}{Method}&Visibility&  Reachability &Success Rate\\  
    & GR& HP & HO & & &\\\hline
       OURS & \ding{51}& \ding{51}&\ding{51}&\textbf{71.7\%}& \textbf{90.2\%}& \textbf{68.5\%}\\ 
       A1 & \ding{55}& \ding{51}&\ding{51}& 69.6\%& 88.0\%& 63.0\%\\  
       A2 & \ding{51}& \ding{51}&\ding{55}& 62.0\%& 77.2\%& 51.1\%\\  
       A3 & \ding{55}& \ding{51}&\ding{55}& 65.2\%& 70.7\%& 50.0\%\\
       A4 & \ding{55}& \ding{55}&\ding{55}& 32.6\%& 00.0\%& 00.0\%
\\ \hline\end{tabular}
    \caption{\textbf{Main Results.} \underline{GR}: Grasp Re-ranking. \underline{HP}: Handover Position estimation. \underline{HO}: Handover Orientation estimation. \S\ref{sec:baseline} explains the implementations for each ablations.} 
    \label{tab:Main Results}
\end{table}

\begin{figure}[h!]
    \centering
    \includegraphics[width=0.9\linewidth]{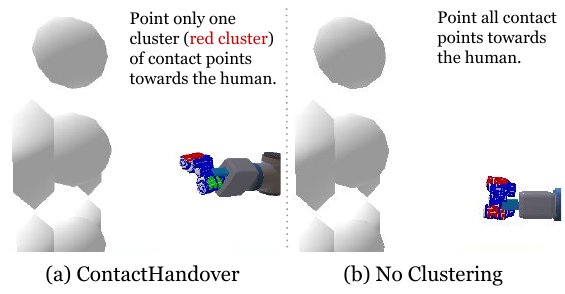}
    \caption{\textbf{Clustering bimodal human contacts.} For objects with bimodal human contact distributions, \name clusters the contact points and only optimizes one cluster. In (a) the robot orients one side of the binoculars towards the human. Without clustering, as in (b), both sides are pointed to the human, leaving neither cluster close to the human.}
    \label{fig:cluster}
\vspace{-2mm}
\end{figure}

\vspace{-2mm}
\section{RESULTS AND ANALYSIS}
\label{sec:result}
\vspace{1mm}
We evaluate \name on a variety of daily objects and compare with several ablations on different components of the system. We show our results quantitatively in Table \ref{tab:Main Results} and qualitatively in Fig \ref{fig:ablation_visuals}.

\textbf{\name achieves the most successful handovers.} As shown in Table \ref{tab:Main Results}, \name achieves visible and reachable handovers for all objects in all runs, with an average success rate of 68.5\%. From our ablations, estimating the handover position (A3) contributes the most to improvements in overall success rate (0\% to 50\%), and in particular, the reachability (0\% to 70\%), compared to no optimization (A4). Ablation 4 does not achieve any successful handovers as all of the objects fall out of reach range of the human arm. 

Our contact-guided grasp selection and handover orientation estimation algorithms further improves handover visibility and reachability. \name improves the final success rate by 18.5 percentage points compared to A3.

\textbf{Grasp re-ranking yields more available human contacts during handover.}
We show the effect of contact-guided grasp selection by comparing with Ablation 1, shown in Fig.~\ref{fig:ablation_visuals}(b).
Without considering human contacts when selecting stable grasps, the robot gripper will often block human contact points on the object. As shown in Fig \ref{fig:ablation_visuals}(b), the robot grasps objects like \textit{hammer} and \textit{pan} by the handles. While it orients the human contact points towards the human during delivery, the robot's gripper blocks human from receiving the objects on the handles. Therefore, the human contacts are less visible due to occlusion from gripper and less reachable due to less contact points between gripper and human.

We note that there are cases where grasp re-ranking makes little effect to the final handover. For example, if all stable robot grasps overlap with human contacts, for instance, on the handle of the knife, the selected robot grasp will block a large portion of human contacts regardless. On the other hand, if the majority of stable robot grasps do not overlap with the predicted human contacts, penalizing human contact occlusion makes little difference. For instance, most stable robot grasps on a mug are on its rim, rather than the handle where human prefers to grasp. 
Nonetheless, qualitatively, on all objects, Grasp re-ranking increases success rate from 63\% (A1) to 68.5\% (\name).

\textbf{Estimating handover orientation faces more human contacts to the receiver.}
We show the effects of handover orientation estimation by comparing with Ablation 2, shown in Fig. \ref{fig:ablation_visuals}(c). In Ablation 2, the robot does not orient contact points towards the human during delivery. 
Although the robot gripper leaves out the human contact parts while grasping, human contact parts are still inaccessible to the human in the final handover. 
For instance, human-preferred contacts could be on the opposite side of the robot gripper from the human, making them unreachable (e.g. \textit{scissors}, \textit{hammer} and \textit{knife} handles); they could also be self-occluded from the human's point of view (e.g. \textit{hammer}, \textit{mug} and \textit{pan} handles). Estimating handover orientation, on average, increases success rate from 51.1\% (A2) to 68.5\% (\name).

\textbf{Clustering human contacts is useful for objects with bimodal contact distributions.}
We show the effect of clustering human contact points in \name compared to no clustering. In an example shown in Fig. \ref{fig:cluster}, \name clusters the human contact points on the binocular and orients the largest cluster (denoted in red) towards the human; in Fig. \ref{fig:cluster}(b), the robot does not cluster the human contact points and estimates the final handover pose by minimizing the distance between all contact points and human. Without focusing on one cluster, the final handover results in a pose where no mode of the human contacts are closer to the human.

\textbf{\name can generalize to unseen objects.}
\label{result:cluster}
We use the YCB dataset\cite{ycb} which is unseen to both the human contact predictor and the robot grasp predictor, and test on both object with comparable classes in ContactDB (hammer and knife) and unseen object classes (spoon, fork, flat screwdriver, and phillips screwdriver). We show the qualitative results in Fig. \ref{fig:ycb}. \name is able to predict human contacts on the handle of these objects. We believe this is because human contact preferences generalize across similar geometries and shapes.

\textbf{Limitations.}
Our work leverages a hand-object contact dataset to learn a proxy for human receiving preferences.
However, there are a few limitations to this approach.
Firstly, our human contact predictions are inferred only from object shape. In certain scenarios, the object state or function may also matter, and cannot be simply inferred from its shape. For example, if a container such as \textit{mug} or \textit{bowl} has water inside, the object should be kept its canonical pose (opening facing up) during the grasping and delivery phase to prevent spilling.
Secondly, our human contact dataset is limited in scale. While the dataset contains 27 diverse household objects, it's still a small fraction of objects we can expect a human or robot to interact with in daily life. Future work could consider expanding the dataset by collecting hand-object contacts for more objects or leveraging multimodal foundation models \cite{delitzas2024scenefun3d,one_shot_affordance_cvpr2024,Laso_2024_CVPR}.
We believe learning from a larger human contact dataset can help increase our method's generalization ability to unseen objects.
Lastly, in our evaluation, we assume a standard human pose in `standing' mode, and that the human receives the object only after the robot delivers the object to the desired pose. Therefore, the policy is not reactive to real-time human pose changes and movements. Future work could consider combining ContactHandover with human pose estimation and hand motion tracking.

\section{CONCLUSIONS}
We propose ContactHandover, a two-phase robot to human handover system guided by human contacts. The robot first achieves a stable grasp on the object while accommodating human contact preferences through a grasp pose re-ranking mechanism. Then the robot delivers the object to a pose that minimizes the human arm joint torques and displacements, as well as the total
distances from contact points to the receiver. To evaluate our system, we propose two quantitative metrics that measure the visibility and reachability of an object's human preferred contacts during handover. We evaluate our system on 27 diverse household objects, demonstrating improved visibility and reachability of predicted human contact areas compared to several ablations.

\addtolength{\textheight}{-0cm}   



\section*{ACKNOWLEDGMENT}
This work was supported in part by the NSF Award \#2037101, and \#2132519. The views and conclusions contained herein are those of the authors and should not be interpreted as necessarily representing the official policies, either expressed or implied, of the sponsors. 

{\small
\bibliographystyle{IEEEtranS}
\bibliography{references}
}
\end{document}